\documentclass[twocolumn, switch]{article} 
\usepackage{times}
\usepackage{epsfig}
\usepackage{graphicx}
\usepackage{amsmath}
\usepackage{amssymb}

\usepackage{multirow}
\usepackage{mathtools}
\usepackage{array}
\usepackage{preprint}

\usepackage{amsmath, amsthm, amssymb, amsfonts}

\usepackage[numbers,square]{natbib}
\usepackage[utf8]{inputenc}	
\usepackage[T1]{fontenc}	
\usepackage{xcolor}		
\usepackage[colorlinks = true,
            linkcolor = purple,
            urlcolor  = blue,
            citecolor = cyan,
            anchorcolor = black]{hyperref}	
\usepackage{booktabs} 		
\usepackage{nicefrac}		
\usepackage{microtype}		
\usepackage{lineno}		
\usepackage{float}			

\usepackage{lipsum}		

\usepackage{newfloat}
\DeclareFloatingEnvironment[name={Supplementary Figure}]{suppfigure}
\usepackage{sidecap}
\sidecaptionvpos{figure}{c}

\usepackage{titlesec}
\titlespacing\section{0pt}{12pt plus 3pt minus 3pt}{1pt plus 1pt minus 1pt}
\titlespacing\subsection{0pt}{10pt plus 3pt minus 3pt}{1pt plus 1pt minus 1pt}
\titlespacing\subsubsection{0pt}{8pt plus 3pt minus 3pt}{1pt plus 1pt minus 1pt}

\title{A Neural Lip-Sync Framework for Synthesizing Photorealistic Virtual News Anchors}

\usepackage{xwatermark}

\newwatermark[firstpage,color=gray!90,angle=0,scale=0.28, xpos=0in,ypos=-5in]{*correspondence: \texttt{zrb915@google.com}}

\usepackage{authblk}

\author[1\thanks{\tt{zrb915@google.com}}]{Ruobing Zheng}
\author[1]{Zhou Zhu}
\author[1]{Bo Song}
\author[1]{Changjiang Ji}
\affil[1]{Deep Innovation R\&D Center, Moviebook, China}

\begin{document}

\twocolumn[ 
  \begin{@twocolumnfalse} 
  
\maketitle

\begin{abstract}
	Lip sync has emerged as a promising technique for generating mouth movements from audio signals. However, synthesizing a high-resolution and photorealistic virtual news anchor is still challenging. Lack of natural appearance, visual consistency, and processing efficiency are the main problems with existing methods. In this paper, we present a novel lip-sync framework specially designed for producing high fidelity virtual news anchors. A pair of Temporal Convolutional Networks are used to learn the cross-modal sequential mapping from audio signals to mouth movements, followed by a neural rendering network that translates the synthetic facial map into high-resolution and photorealistic appearance. This fully-trainable framework provides an end-to-end processing that outperforms traditional graphics-based methods in many low-delay applications. Experiments also show the framework has advantages over modern neural-based methods in both visual appearance and efficiency.
\end{abstract}
\vspace{0.35cm}

  \end{@twocolumnfalse} 
] 



\section{Introduction}

Virtual news anchor is an emerging application that landing artificial intelligence technology in the field of news media. Researchers aim to build a video-based news anchor that could automatically broadcast the given news, with the photorealistic appearance and person-specific speaking style. Such virtual news anchors can be on call any time to deal with breaking news, keeping tireless and stability all day long. 

Technically, techniques related to virtual anchors have been widely studied in the literature of computer vision and graphics, including audio or text-driven lip sync \cite{wang2011text,fan2015photo,suwajanakorn2017synthesizing}, 3D model-based facial animation \cite{cudeiro2019capture,karras2017audio,tian2019audio2face,fried2019text}, and some GAN-based talking face solutions \cite{zhou2019talking,vougioukas2019end,zakharov2019few}. The main idea is to learn the cross-modality relationship between audio or text signals to the visual information of a target person. However, it is still challenging to produce a realistic virtual anchor with current methods\cite{wang2019neural}.

There are two main problems in applying current methods to the virtual anchor projects. The lack of video resolution, visual consistency and natural appearance is the first obstacle. Some few-shot studies successfully synthesize a talking face via one \cite{chung2017you,vougioukas2019end} or a few images \cite{zakharov2019few}. However, their results lack sufficient resolution and details when applying to real-world applications. A series of methods \cite{cudeiro2019capture,xiao2018dense,yi2020audio} use 3D Morphable Model (3DMM) as an intermediate representation to parameterize facial expressions and mouth movements. However, registering and rendering 3DMM is also a heavy task \cite{thies2019neural,fried2019text}, and personal talking styles may be lost and diluted. Secondly, the lack of processing efficiency also prevents current methods from low-delay scenarios. Several lip-sync studies achieve high-quality performance due to they only ``rewrite"  the mouth region in template faces. But such a strategy requires time-consuming steps like candidate frame selection in past graphics-based methods \cite{suwajanakorn2017synthesizing,wang2011text}. A recent study \cite{kumar2017obamanet} presents a fully-trainable lip-sync solution. Its neural rendering model improves the processing efficiency but also brings visible mismatches between synthesized mouths and original faces.

To tackle the above issues, we propose a novel lip-sync framework specially designed for synthesizing high-resolution and photorealistic virtual news anchors. We bring the advanced neural architectures and rendering strategy in the traditional two-stage lip-sync scheme \cite{suwajanakorn2017synthesizing}. This design is to keep promising performance while improving processing efficiency via deep learning techniques. As for driving signals, we use audio input because it is easy to generate from text and has better alignment with mouth movements. In summary, the contributions of this paper include the following three points:

\begin{itemize}
	\item  We model the cross-modal sequential mapping from audio signals to mouth movements by a pair of adversarial Temporal Convolutional Networks. Experiments show our model outperforms traditional RNN-based baselines in both accuracy and speed.  
	
	\item We propose a neural rendering solution for producing high-resolution and photorealistic appearance from synthetic facial maps. The image-to-image translation-based method not only avoids the cumbersome operations in traditional graphics-based methods but also solves the existing issues in recent neural-based solutions.
	
	\item We comprehensively evaluate our lip-sync framework at both the audio-to-mouth and rendering stages. We also demonstrate its application for building a virtual news anchor. The experience is valuable for related applications and studies.
	
\end{itemize}
\begin{figure*}[tb!]
	\centering
	\includegraphics[width=1\textwidth]{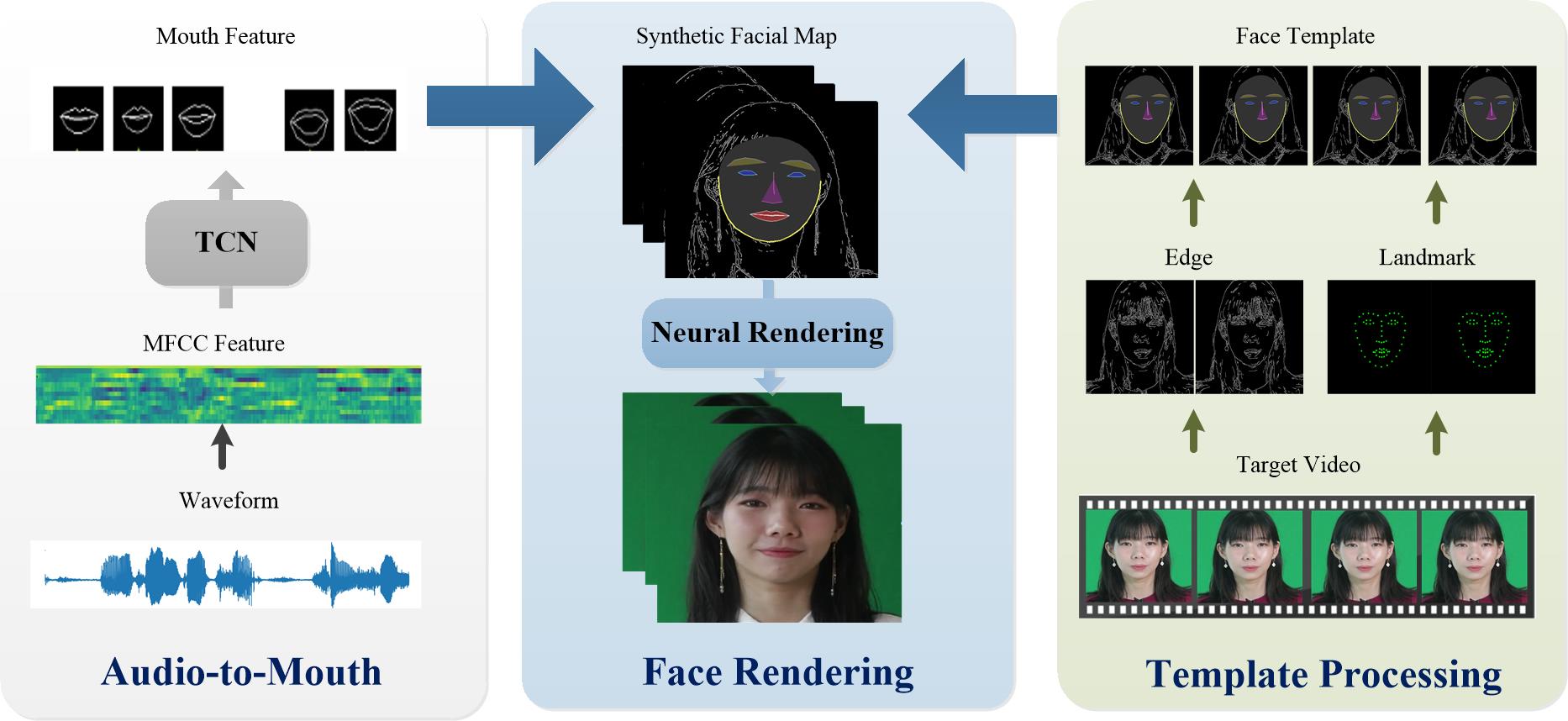}
	\caption{Flowchart of the proposed neural lip-sync framework. The TCN-based mapping network generates time-series mouth features from an audio clip. The generated mouths are integrated into the template face and then rendered by the neural rendering network.
	}
\end{figure*}

\section{Related Work}

\subsection{Lip Sync}

Lip-sync studies \cite{bregler1997video,busso2007rigid} focus on generating realistic human-speaking videos with accurate lip movements, based on the given speech content and a target video clip. This topic has been widely explored in computer graphics literature. The major solution is to learn a mapping from audio features to visual features and then render them into a photorealistic texture \cite{suwajanakorn2017synthesizing,taylor2017deep}. Earlier methods \cite{wang2011text,xie2007coupled} use probabilistic graphical models, represented by Hidden Markov Models, to capture the correspondences between mouth movement and audio. As for rendering, the graphics-based way is to select the best matching frames from a set of candidates \cite{suwajanakorn2017synthesizing}. Recently, deep learning practitioners have successfully applied neural models in both audio-to-mouth and rendering stages. A litany of studies \cite{fan2015photo,tian2019audio2face} employ Recurrent Neural Network (RNN), such as Long Short-Term Memory (LSTM) and Bidirectional LSTM, to learn the audio-to-mouth mapping. A recent work \cite{kumar2017obamanet} shows the power of image-to-image translation models in synthesizing the photorealistic facial appearance. Compared to graphics-based rendering methods, deep learning technologies show the advantage in speed but still have visible flaws in current solutions.

\subsection{Temporal Convolutional Networks}
RNNs are once the standard choice in modeling sequence problems. However, some significant issues still limit RNN-based models, including vanishing gradients and memory-bandwidth limitation \cite{martens2011learning}. Although some successful derivatives \cite{gers1999learning} relieve the problems, they still hardly achieve satisfactory performance on long sequences. Recently, hierarchical models show more power in learning sequential correspondence \cite{culurciello2018fall}. A representative work is Temporal Convolutional Network (TCN) \cite{bai2018empirical} which distills best prior practices in convolutional network design into a convenient but powerful convolutional sequential model, such as dilated convolutions, causal convolutions, and residual connections. A comprehensive experiment \cite{bai2018empirical} demonstrates that TCN outperforms canonical recurrent networks across a diverse range of tasks and datasets.

\section{Proposed Method}

Our lip-sync framework can be interpreted as two stages of work: a pair of Temporal Convolutional Networks (TCN) learning the seq-to-seq mapping from audio to lip movements, and an image-to-image translation-based renderer generating high-resolution and photoreal texture from intermediate face representation. We illustrate the flowchart in Figure 1.

\begin{figure*}[tb]
	\centering
	\includegraphics[width=0.95\textwidth]{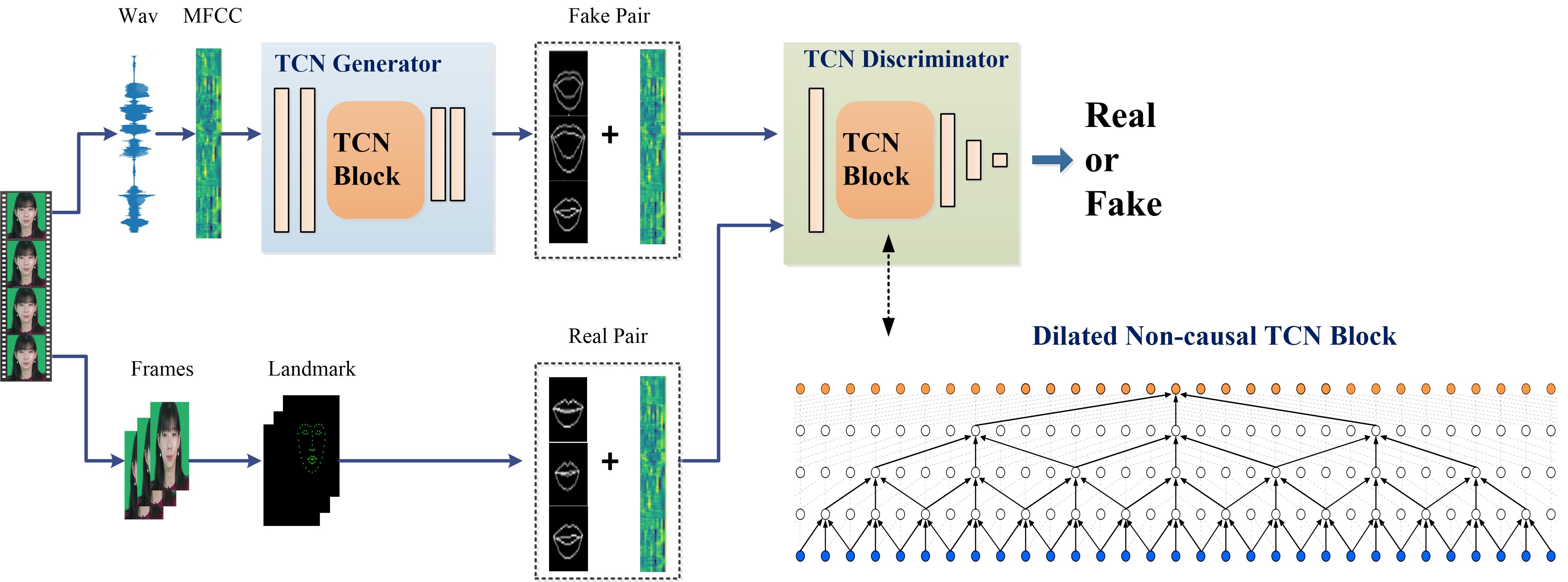}
	\caption{The architecture of the TCN-based adversarial model used in the audio-to-mouth mapping. The TCN-based generator learns the mapping from MFCC features to time-series mouth features. Another TCN-based discriminator takes the combination of audio and mouth features as input and judges the real and fake (synthesized) pairs.}
\end{figure*}

\subsection{Audio-to-Mouth Mapping}

\subsubsection{Adversarial Temporal Convolutional Networks}

Different from most RNN-based implementations \cite{kumar2017obamanet,xiao2018dense,tian2019audio2face}, we employ TCN to learn the audio-to-mouth mapping. An empirical study \cite{bai2018empirical} has shown that TCN outperforms generic RNN-based models on a variety of sequence modeling tasks, with the strength of the large sequential perceptive field, stable gradients, and low memory requirements.

To bring such strengths into the lip-sync task, we tailor a pair of TCN-based model in an adversarial framework, as shown in Figure 2. A TCN-based generator learns the mapping from audio features to mouth features. It consists of four 1-D convolutions layers, two fully-connected layers, and a TCN block. The TCN block is wrapped with 1-D convolutions layers which downsample the audio features (100 fps) to the video rate (25 fps). The generator accepts a 200 length audio sequence and outputs a 10-D vector on each node of a 50 length sequence. We also devise a similar TCN-based discriminator to support the training of the mapping network. The discriminator takes the combination of audio and mouth sequences as input and outputs a real or fake label. 

Within the TCN block, dilated convolutions \cite{oord2016wavenet} are used to receive the exponentially large reception field on the long sequence. According to previous reports \cite{suwajanakorn2017synthesizing,fan2015photo}, lip motion depends on both the past and future audio signals. Hence, we employ the non-causal structure in our TCN block to consider both future and past signals. All TCN blocks used in our networks are with the kernel size 3, 
256 filters, and dilation factor [1,2,4,8], as shown in Figure 2. The implementation and illustration of TCN block are adopted from the work \cite{Philippe2018}.


\subsubsection{Loss Function}

The loss consists of an $ L_{2} $ regression loss, a pairwise inter-frame loss, and an adversarial loss. We define that the model G as the mapping of audio-to-mouth pairs, which can be  denoted as $ \left\{\left(\mathbf{a}_{i}, \mathbf{m}_{i}\right)\right\}_{i=1}^{F} $, where $\mathbf{a}_{i} \in \mathbb{R}^{200 \times 13}$ and $\mathbf{m}_{i} \in \mathbb{R}^{50 \times 10}$. Therefore, we use $G(\mathbf{a_{i}})\in \mathbb{R}^{50 \times 10}$ to represent the predicted mouth features.

The $ L_{2} $ regression loss measures the mapping accuracy on individual features:

\begin{equation}
\mathcal{L}_{L_{2}}(G)=\left\|\mathbf{m}_{i}-G(\mathbf{a}_{i})\right\|_{F}^{2}
\end{equation}

The pairwise inter-frame loss computes the $ L_{2} $ distance between the differences of consecutive frames between predicted feature sequences and target sequences, which is used to increase the temporal stability \cite{cudeiro2019capture}: 

\begin{equation}
\mathcal{L}_{int}(G)=\left\|\left(\mathbf{m}_{i}-\mathbf{m}_{i-1}\right)-\left(G(\mathbf{a}_{i})-G(\mathbf{a}_{i-1})\right)\right\|_{F}^{2}
\end{equation}

Moreover, the adversarial loss is used with the TCN-based discriminator to capture high-level discrepancy between generated and target features, which can be defined as:

\begin{equation}
\begin{split}
\mathcal{L}_{GAN}(G, D)= & \mathbb{E}_{\mathbf{m},\mathbf{a}}[\log D(\mathbf{m},\mathbf{a})]+ 
\\&\mathbb{E}_{\mathbf{m},\mathbf{a}}[\log (1-D(G(\mathbf{a}),\mathbf{a}))]
\end{split}
\end{equation}

Our final objective ($ \lambda_{1} =100, \lambda_{2} =1$) is defined as  :

\begin{equation}
G^{*} = \min _{G} \max _{D} \mathcal{L}_{G A N}(G, D)+\lambda_{1} \mathcal{L}_{L_{2}}(G)+\lambda_{2} \mathcal{L}_{int}(G)
\end{equation}

\subsubsection{Overlapping between Consecutive Sequences}

TCN uses paddings to keep the same length between input and output sequences. However, padding signals impact the predictions in both the head and tail of the output sequence. To avoid this issue, we create overlaps on both input and output sequences. The overlapping range is related to the TCN's receptive field, and we consider it as a hyperparameter and determined by later experiments.

\subsection{Neural Face Rendering}

Inspired by Kumar \cite{kumar2017obamanet}, we devise a neural rendering module based on the pix2pixHD, a hierarchical image-to-image translation model \cite{wang2018high}. 
We first synthesized the specially designed facial maps as an intermediate representation. Then these facial maps are sent to pix2pixHD to generate high-resolution (512 $\times$ 512) face appearance. This fully-trainable solution not only avoids the time-consuming operations in past graphics-based methods \cite{wang2010synthesizing,suwajanakorn2017synthesizing} but also solve existing issues in the recent study \cite{kumar2017obamanet}. We will compare them in later experiments. 

\subsubsection{Synthetic Facial Maps}

As most lip-sync studies \cite{kumar2017obamanet,suwajanakorn2017synthesizing}, we integrate the generated mouth into a face template. We design a synthetic facial map as the intermediate representation to bridge the gap between generated mouths and final face appearance. We first extract the Canny edges and facial landmarks from each frame in the template video. Then, we recover the generated mouth features into the template face and fine-tune the jawline. Finally, we draw the facial maps based on the composite face landmark and background edges, as shown in Figure 3.

\begin{figure}[tb!]
	\centering
	\includegraphics[width=1\columnwidth]{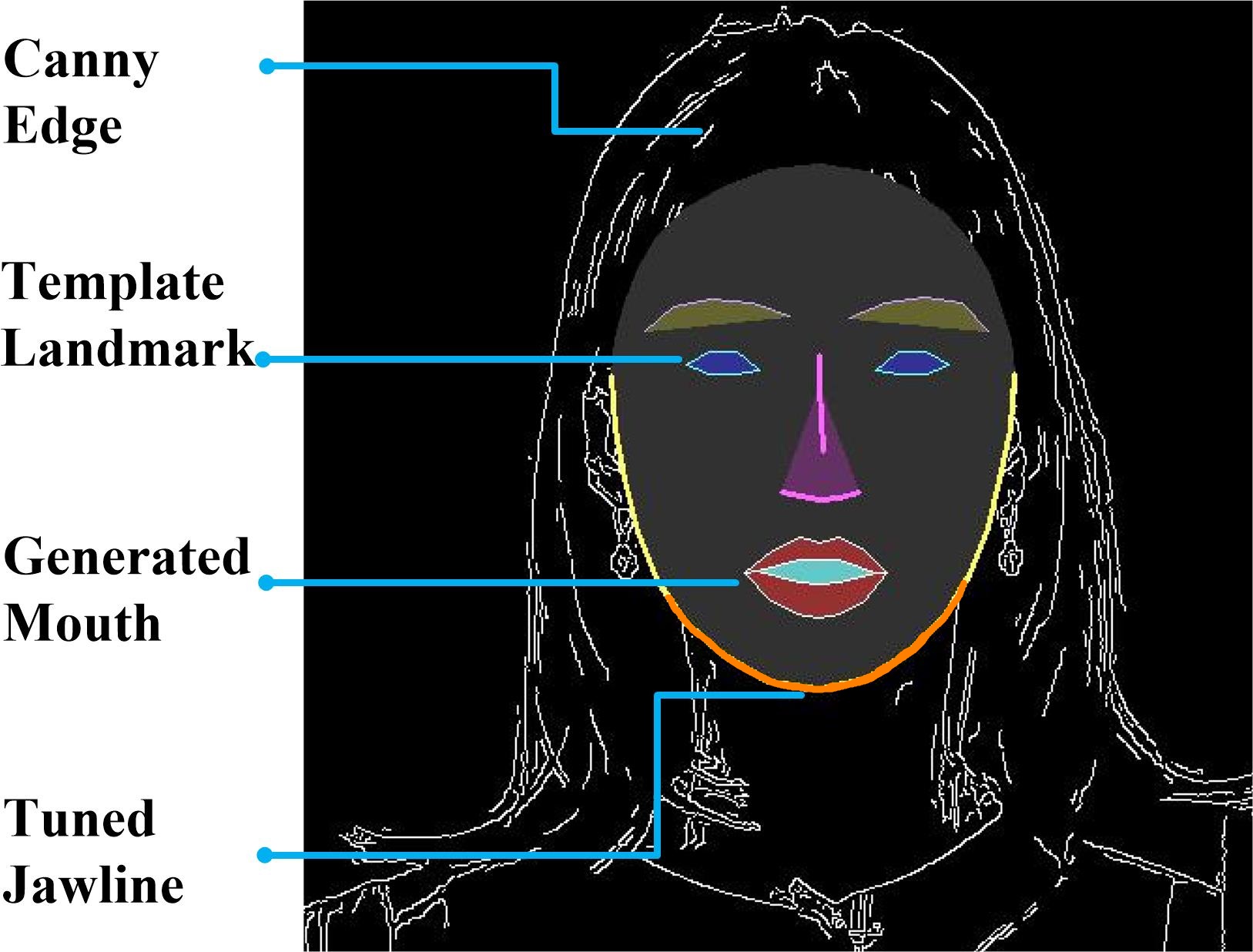}
	\caption{The synthetic facial map consists of the Canny edge and facial landmark from the template frame, the generated mouth shape, and the tuned jawline.}
	\label{fig:graph}
\end{figure}

Generating continuous and accurate details is one of the main challenges for current rendering methods, especially for generating high-resolution videos. Previous rendering methods \cite{wang2018video,chan2019everybody} use the optical flow or temporal-consistent losses to improve visual consistency. Our solution is to directly provide necessary information via Canny edges. Later experiments show it effectively improves the details and inter-frame consistency of the parts like hair, earrings, and clothes.

\subsubsection{Jaw Correction}

Jaw correction is a necessary step in past graphic-based studies \cite{mcallister1997lip,suwajanakorn2017synthesizing}. In our image-to-image translation solution, the jaw is also important because the encoder-decoder structure integrates the jaw and mouth into high-level features and then interprets it globally. Figure 4 illustrates how different jawlines affect the rendered results with the same mouth shape. This result confirms that simply merging the generated mouth into a talking face template will induce the rendering error. Considering the rigid relationship between the jawline and mouth shape \cite{jiang2002relationship}, we use a simple multi-linear regression model to learn how the mouth shape (w,h) affects the offset of the jawline, as shown in Figure 4. To cooperate with the above strategy, we specifically select a non-speaking video clip as the template video and dynamically tune the jawline with the generated mouth shapes.

\begin{figure}[tb!]
	\centering
	\includegraphics[width=0.85\columnwidth]{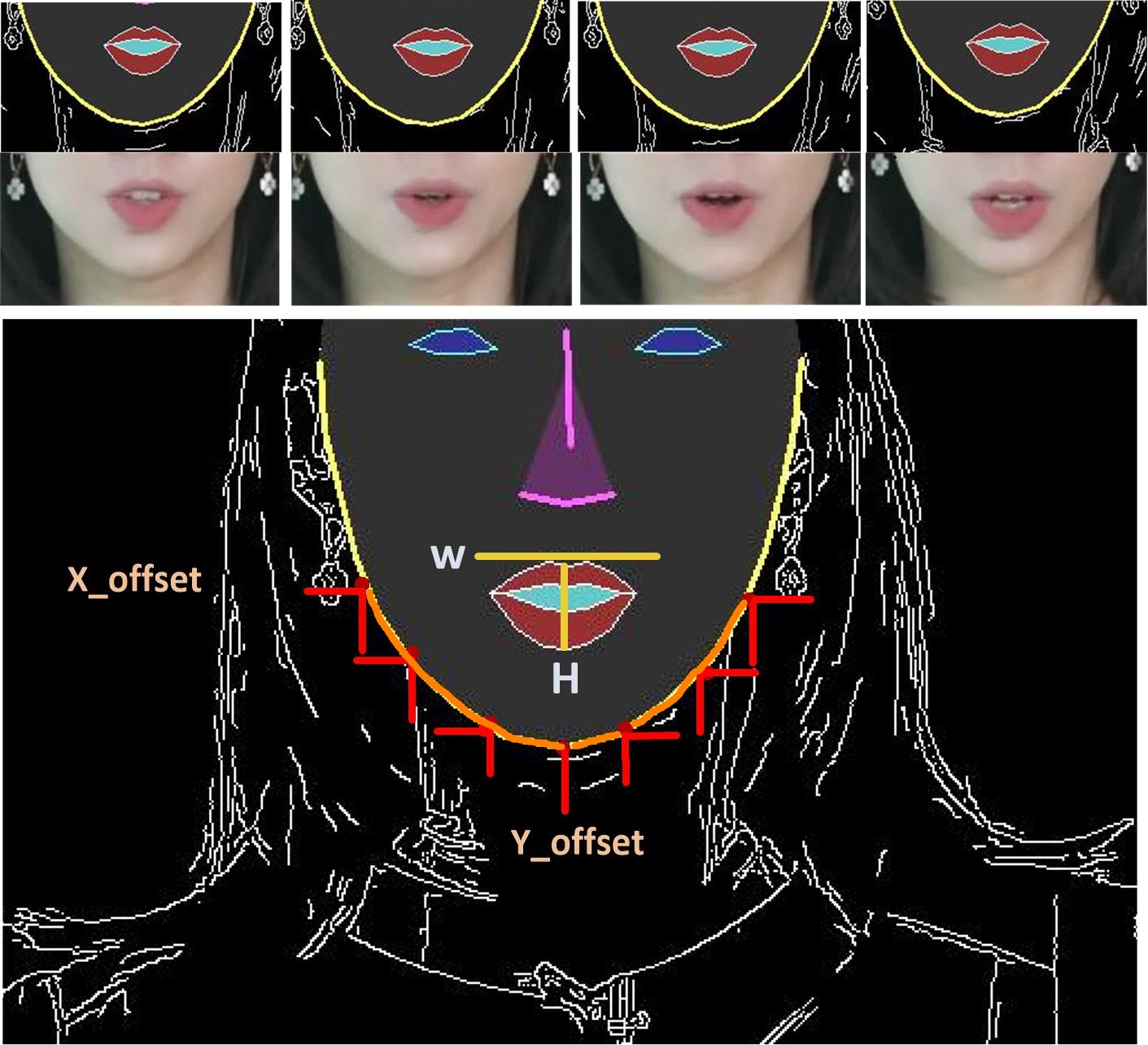}
	\caption{The upper part shows the different results rendered from the same mouth shape and different jawlines. The lower part illustrate our jaw correction strategy that dynamically tunes the jawline based on the generated mouth shapes.}
	\label{fig:graph}
\end{figure}

\section{Experiments}

In this section, we evaluate our lip-sync framework at both the audio-to-mouth mapping and the rendering stages. We also demonstrate the process of building a virtual news anchor for a target person. All neural networks are implemented in Keras 2.2.4 and Tensorflow 1.14.0. We processed the data on Inter Xeon silver 4110 and trained neural models on 4 NVIDIA Titan V.

\subsection{Dataset}

To cooperate with the lip-sync framework, we recorded 2-hour training videos from a female actor as the news anchor manner. The actor was required to read shot scripts (5-15 min) from a teleprompter in the green screen environment. The actors are asked to read at different speeds and avoid out-plane head rotation. The resolution of recorded videos is 1920 $\times$ 1080, and the face area is 512 $\times$ 512. We shot the videos under the stable lighting condition. Both the mouth features and audio features were extracted from the video at the 25 fps and 100 fps, respectively. 

\subsection{Feature Processing}
We extract 68 \emph{dlib} facial landmarks from each frame in training videos. We get the 20 normalized mouth coordinates via removing the in-plane rotation and calculating the relative coordinates based on the center of the nose. We use PCA to de-correlate the mouth coordinates into a 10-D feature vector because the first 10 principal components cover nearly 99\% variability in original mouth landmarks. As for audio features, we use the handcrafted acoustic feature, Mel-frequency cepstral coefficients (MFCC)  as previous studies \cite{tian2019audio2face,cudeiro2019capture}. Such time-frequency representations are better aligned with human auditory perception \cite{vasquez2019melnet}. The sampling rate of 13-D MFCC features is 100 Hz.

\begin{table}[tb!] 
	\centering
	\caption{Comparing the performance of audio-to-mouth mapping between the proposed model and baselines.}
	\linespread{1.5}\selectfont
	\begin{tabular}{m{0.3\columnwidth}m{0.14\columnwidth}<{\centering}m{0.14\columnwidth}<{\centering}m{0.15\columnwidth}<{\centering}}   
		\hline 
		Model & MSE & MAE  & Int-MSE    \\ 
		\hline 
		Time-delayed LSTM & 0.00366  &  0.0465 & 0.00735 
		\\ 
		Bi-LSTM & 0.00357 & 0.0458 & 0.00712
		\\ 
		
		Non-Causal TCN  & 0.00155 & 0.0278  & \textbf{0.00122} 
		\\

		Adversarial TCN (our)  & \textbf{0.00141} & \textbf{0.0261 } & 0.00132 \\ 
		\hline   
		
	\end{tabular}  
	\label{table1}
\end{table}

\subsection{Audio-to-Mouth Evaluation}

\subsubsection{Metrics}

At this stage, the models accept time-series audio features and output PCA mouth features. To fairly compare the audio-to-shape mapping performance between our model and baselines, besides two common regression metrics (MSE and MAE), we employ the Inter-Frame MSE which measures the frame-wise velocity, which can be written as:

\begin{equation}
\mathrm{MSE_{int}}=\frac{1}{n}\sum_{i=1}^{n}\left\| \left(Y_{i}-Y_{i-1}\right) - \left(  \hat{Y}_{i}-\hat{Y}_{i-1}\right) \right\|^{2}_{2}
\end{equation}

\subsubsection{Baselines}

We select two typical RNN baselines from recent lip-sync studies. We also compare the proposed model with a basic TCN generator.

\begin{itemize}
	\item \textbf{Time-delayed LSTM} is a typical RNN-based implementation for learning the audio-to-mouth mapping. Suwajanakorn \cite{suwajanakorn2017synthesizing} reports that the time delay mechanism is effective to consider future audio signals. Here we implement this model based on the open-source code from the recent work \cite{kumar2017obamanet}.
	
	\item \textbf{Bidirectional LSTM} is also a popular choice in recent speech recognition and facial animation studies \cite{graves2005framewise,tian2019audio2face}. Similar as time-delay LSTM, the bidirectional architecture computes both forward state sequence and backward state sequence.
	
	\item \textbf{Non-Causal TCN} covers both past and future signals with non-causal convolutions. This baseline is actually the generator of our TCN-based adversarial network. We train it independently with the same $L_{2}$ loss as RNN baselines.

\end{itemize}	

\subsubsection{Model Performance}

As shown in Table 1, we observe that TCN-based models significantly outperform RNN baselines on all three metrics. Between RNN-based models, the bidirectional structure brings a slight improvement than the time-delayed LSTM. We also notice that the adversarial TCN has taken a step forward than non-causal TCN. Augmenting the TCN architecture with the adversarial loss successfully reduces the MSE and MAE. It confirms that the TCN discriminator captures the high-level discrepancy and facilitate the training process. 

Next, we compare the training time and inference time of LSTM, Bidirectional LSTM, and TCN architecture. For a fair comparison, we use the same $L_{2}$  loss as LSTMs to train the TCN generator (Non-causal TCN). All test models are trained with the same batch-size using one NVIDIA Titan V. For inference time, we process a one-minute audio clip and record the processing times. As shown in Table 2, TCN significantly outperforms RNN-based models at both training and inference stage. This result confirms that the TCN architectures are capable of improving the processing efficiency in lip-sync tasks. This advantage will benefit many real-time applications.

\begin{table}[tb!] 
	\centering
	\caption{Comparing the training and inference time (1-min audio) between LSTM, Bidirectional LSTM, and TCN.}
	\linespread{1.5}\selectfont
	\begin{tabular}{m{0.15\columnwidth}m{0.2\columnwidth}<{\centering}m{0.2\columnwidth}<{\centering}<{\centering}m{0.2\columnwidth}<{\centering}}
		
		\hline 
		Models & Batch training (s) & Total training (min) & Inference time (s) \\
		\hline

		LSTM & 0.069 $ \pm $ 0.005 & 67.43 $ \pm $ 5.62 & 2.272 $ \pm $ 0.269   \\
		
		Bi-LSTM & 0.124  $ \pm $ 0.007  & 114.58  $ \pm $ 3.76  &3.376 $ \pm $ 0.201  \\
		
		TCN & \textbf{0.068  $ \pm $ 0.005 }  &\textbf{35.82 $ \pm $ 2.62 }  & \textbf{0.011 $ \pm $  0.005} \\
		\hline
		
	\end{tabular}  
	\label{table1}
\end{table}

\subsubsection{Overlapping Range}

To support the aforementioned overlapping strategy, we examine the frame-wise accuracy within the 50 length output sequence. We calculate the MSE on each frame and draw in the Figure 5.We observe that the errors of the first four frames and the last eight frames are significantly larger than others. It indicates that these frames are influenced by the padding signals from TCN architectures. Figure 5 also indicates that the mouth movements are more dependent on future audio signals. This finding contradicts some previous studies \cite{kumar2017obamanet,suwajanakorn2017synthesizing} that use shorter future audio signals than the past in LSTMs. But we have not verified it due to the symmetrical structure of non-causal TCN. According to this result, we overlap the output sequence by 10 frames on both head and tail, and the corresponding input sequence is overlapped by 40 frames.

\begin{figure}[tb]
	\centering
	\includegraphics[width=0.85\columnwidth]{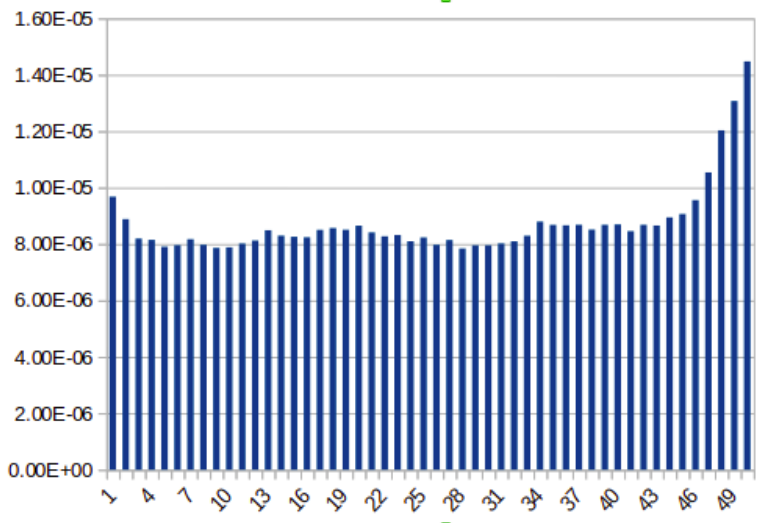}
	\caption{We plot the MSE on each frame in the 50-length output sequence. The errors of the head and tail frames are significantly larger than others.}
	\label{fig:graph}
\end{figure}

\subsection{Rendering Evaluation}

So far, few open-source lip-sync methods can generate high-resolution face texture, especially in the same two-stage framework. For a fair comparison, we choose the most similar method  \cite{kumar2017obamanet} as the baseline, which also applies image-to-image translation to the rendering stage. We experiment on both the Obama dataset and our greenscreen dataset.

The baseline method uses pix2pix to rewrite the mouth region based on newly generated skeletons. Figure 6 shows some defects caused by this strategy. In left Obama images, we observe an obtrusive patch area where the texture does not match the surroundings. For the right greenscreen images, we replace the pix2pix \cite{isola2016image} with advanced pix2pixHD \cite{wang2018high} but still keep the same inpainting strategy. We find the patch issue has been alleviated, but defects still exist, such as multi-layered teeth and noise pixels. We review source videos and believe the patch problem is caused by the inconsistent lighting condition in the Obama dataset, and the inpainting strategy magnifies this issue. In the right images, the baseline rendering method shows other types of defects, and even pix2pixHD can not make the remedy.

Different from the baseline, we use the synthetic facial map as the intermediate representation. Except for facial landmarks, we extract Canny edges from template frames to maintain the details and visual consistency of the parts like hair, earrings, and clothes. Figure 7 confirms that the lack of edge information leads to inaccurate details in rendered results.

\begin{figure}[tb!]
	\centering
	\includegraphics[width=0.95\columnwidth]{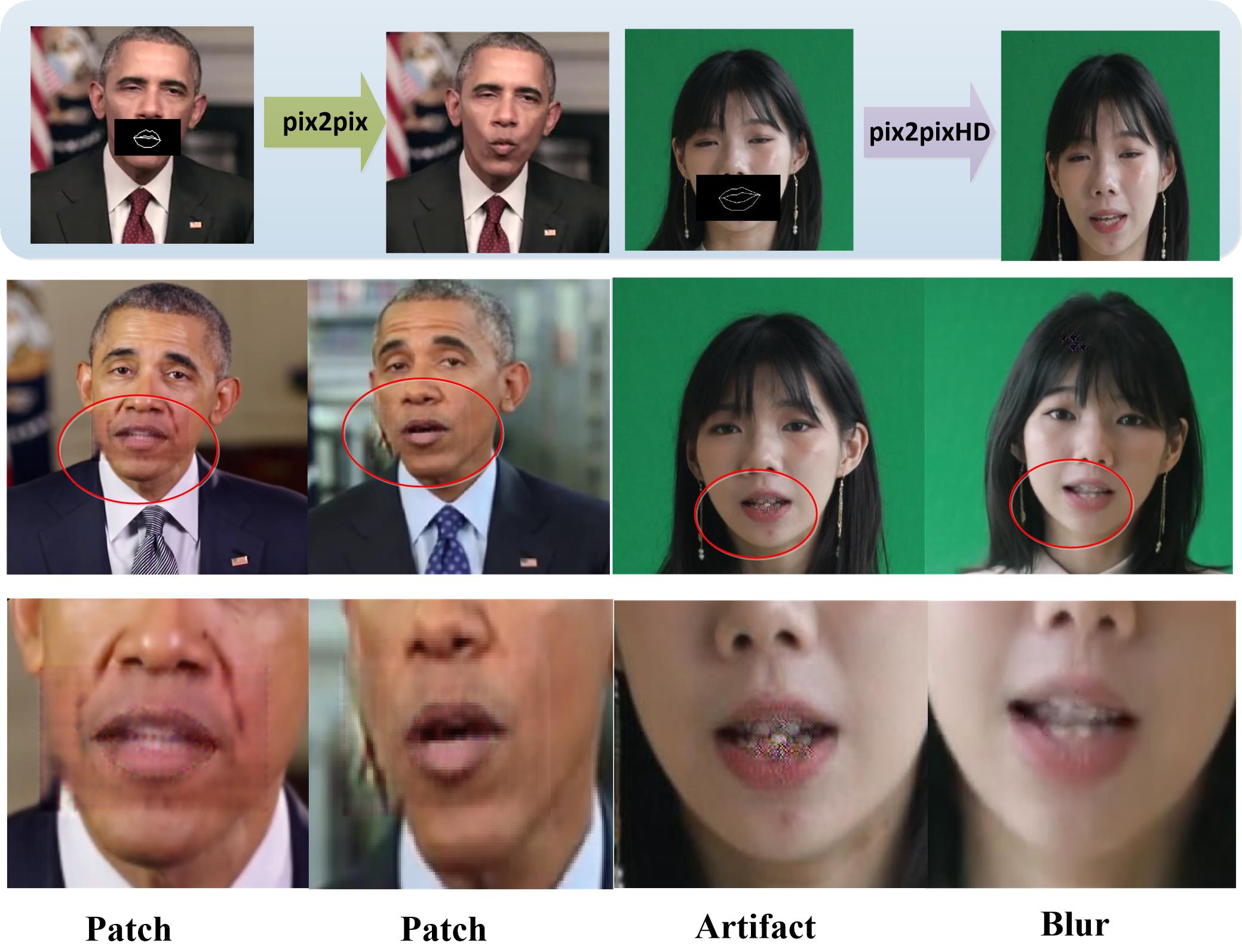}
	\caption{We illustrate some defects caused by the baseline rendering method. Both the Obama and greenscreen images are rendered based on the inpainting strategy from the baseline study.}
	\label{fig:graph}
\end{figure}

\begin{figure}[tb!]
	\centering
	\includegraphics[width=1\columnwidth]{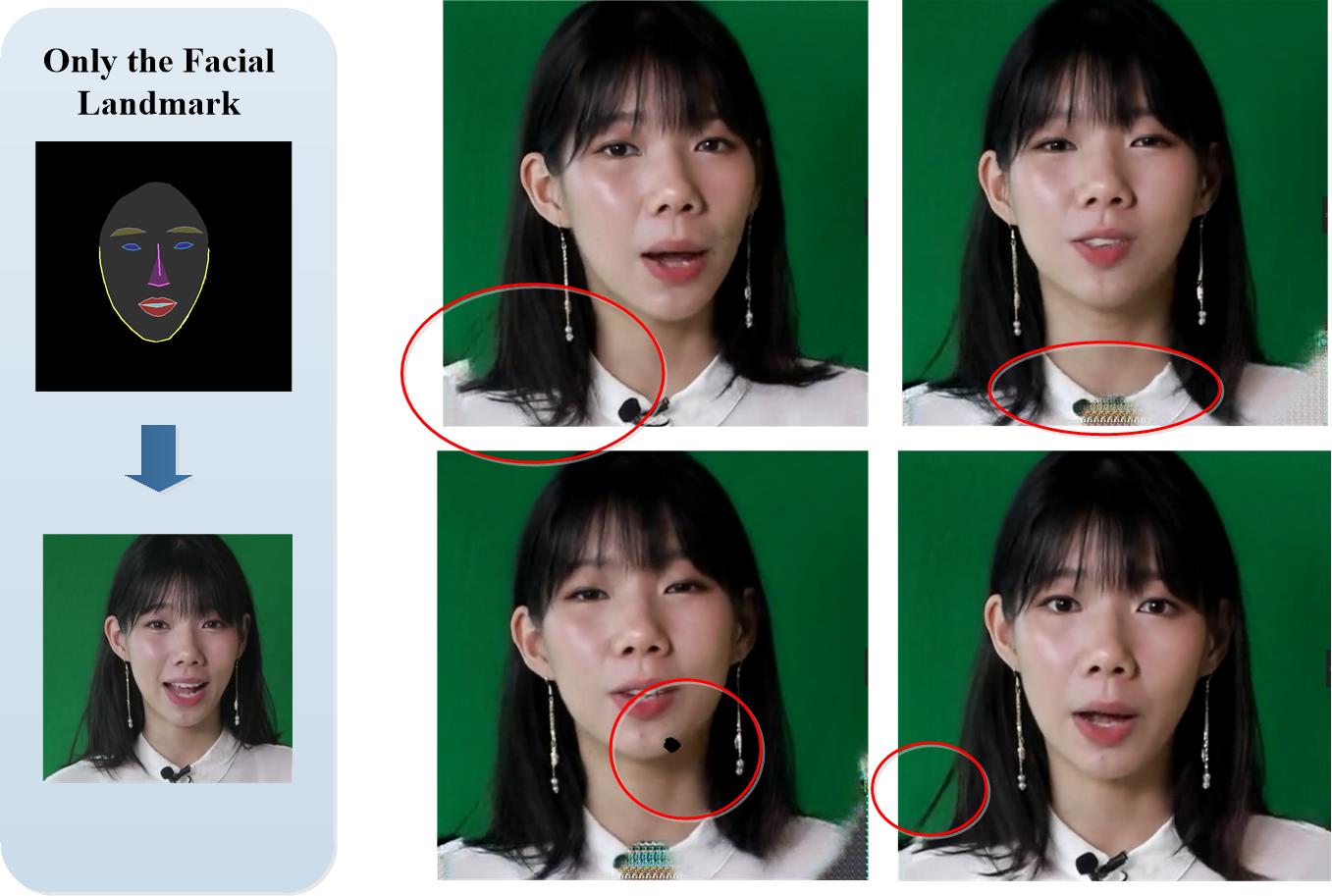}
	\caption{The lack of edge information leads to inaccurate details in rendered results.}
	\label{fig:graph}
\end{figure}

\begin{figure*}[tb!]
	\centering
	\includegraphics[width=.95\textwidth]{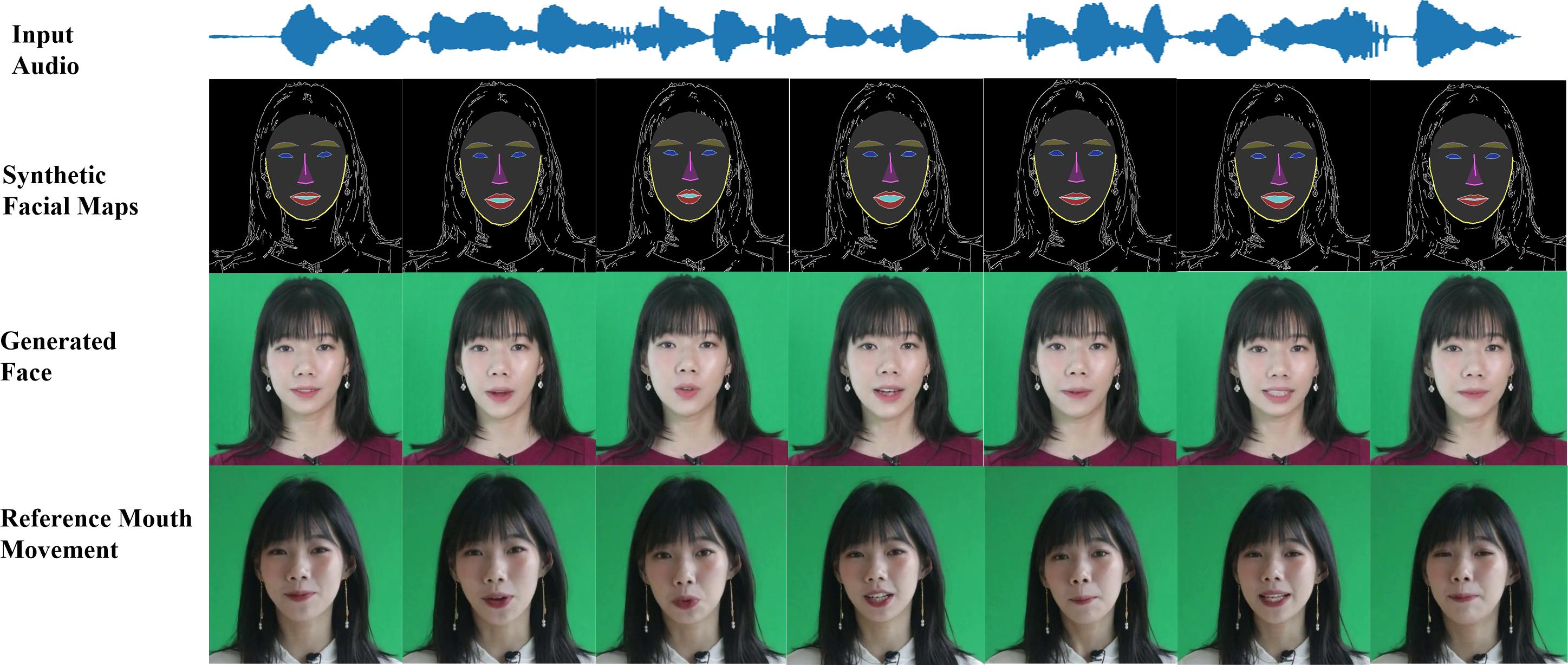}
	\caption{The generated 512 $ \times $ 512 faces using our neural rendering method. From the bottom upward: the reference mouth movements in the audio-source video, the generated faces, and the synthetic facial map used as intermediate representation.}
\end{figure*}

Figure 8 shows the generated 512 $ \times $ 512 video frames using our neural rendering method. Considering there is no direct ground truth due to we generate new mouth movement on a fixed video clip, we provide the intermediate facial maps and the mouth movements in the audio-source video as a reference. Our results show good visual compatibility and embouchure consistency. We observe that the rendered results accurately capture the mouth movements in the sound-source video while representing realistic facial expressions. However, our method still shows some limitations. For some ``big" embouchures, the synthesized mouths show less sensitivity. The lower teeth in generated frames seem to be blurrier than upper teeth (column 2), and the gap between upper and lower teeth has not been fully recovered (column 6).

As shown in Figure 9, the generated talking face is integrated with a pre-recorded body template. We remove the greenscreen background and create a virtual studio environment to work with the virtual news anchor. We provide a demo video in our supplementary material.

\begin{figure*}[tb!]
	\centering
	\includegraphics[width=.95\textwidth]{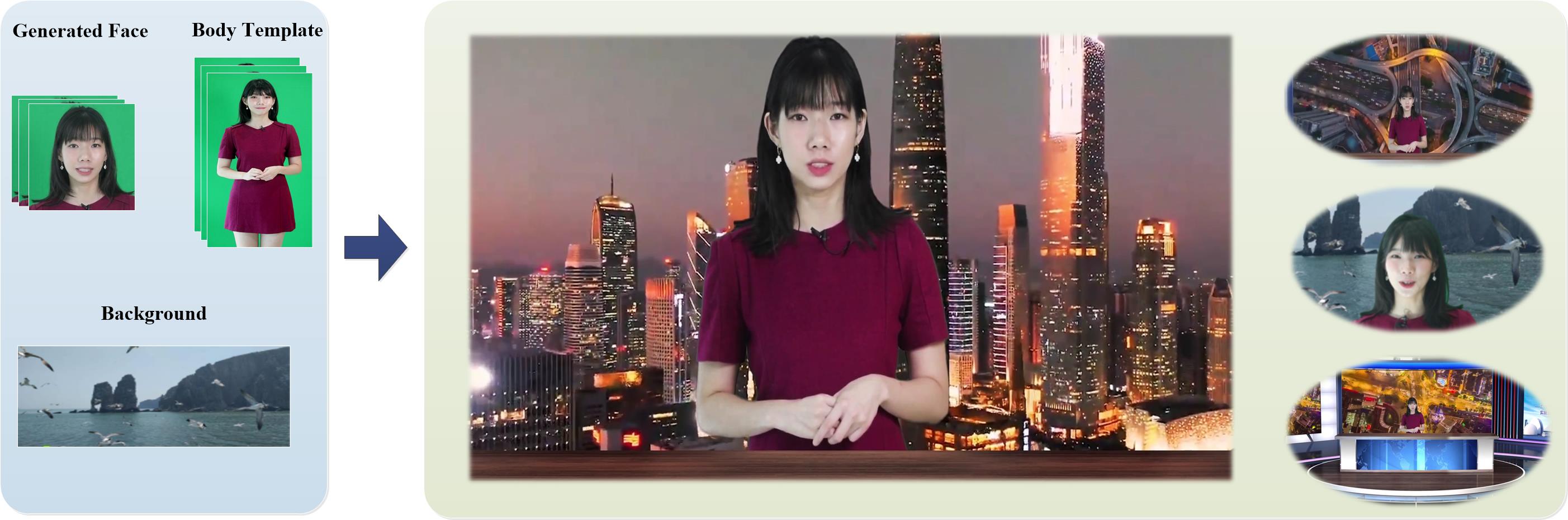}
	\caption{Combining the generated face, body template and background to synthesize the final virtual anchor.}
\end{figure*}

\section{Conclusion}
This paper describes a novel lip-sync framework and demonstrates its application in synthesizing high-resolution and photorealistic virtual news anchor. We augment the traditional two-stage lip-sync framework with advanced deep learning architectures and the compatible rendering strategy, effectively improving both visual appearance and processing efficiency. Our work will inspire more valuable studies and benefit related applications.

\normalsize
\bibliography{egbib}

\begin{thebibliography}{35}
\providecommand{\natexlab}[1]{#1}
\providecommand{\url}[1]{\texttt{#1}}
\expandafter\ifx\csname urlstyle\endcsname\relax
  \providecommand{\doi}[1]{doi: #1}\else
  \providecommand{\doi}{doi: \begingroup \urlstyle{rm}\Url}\fi

\bibitem[Wang et~al.(2011)Wang, Han, Soong, and Huo]{wang2011text}
Lijuan Wang, Wei Han, Frank~K Soong, and Qiang Huo.
\newblock Text driven 3d photo-realistic talking head.
\newblock In \emph{Twelfth Annual Conference of the International Speech
  Communication Association}, 2011.

\bibitem[Fan et~al.(2015)Fan, Wang, Soong, and Xie]{fan2015photo}
Bo~Fan, Lijuan Wang, Frank~K Soong, and Lei Xie.
\newblock Photo-real talking head with deep bidirectional lstm.
\newblock In \emph{2015 IEEE International Conference on Acoustics, Speech and
  Signal Processing (ICASSP)}, pages 4884--4888. IEEE, 2015.

\bibitem[Suwajanakorn et~al.(2017)Suwajanakorn, Seitz, and
  Kemelmacher-Shlizerman]{suwajanakorn2017synthesizing}
Supasorn Suwajanakorn, Steven~M Seitz, and Ira Kemelmacher-Shlizerman.
\newblock Synthesizing obama: learning lip sync from audio.
\newblock \emph{ACM Transactions on Graphics (TOG)}, 36\penalty0 (4):\penalty0
  95, 2017.

\bibitem[Cudeiro et~al.(2019)Cudeiro, Bolkart, Laidlaw, Ranjan, and
  Black]{cudeiro2019capture}
Daniel Cudeiro, Timo Bolkart, Cassidy Laidlaw, Anurag Ranjan, and Michael~J
  Black.
\newblock Capture, learning, and synthesis of 3d speaking styles.
\newblock In \emph{Proceedings of the IEEE Conference on Computer Vision and
  Pattern Recognition}, pages 10101--10111, 2019.

\bibitem[Karras et~al.(2017)Karras, Aila, Laine, Herva, and
  Lehtinen]{karras2017audio}
Tero Karras, Timo Aila, Samuli Laine, Antti Herva, and Jaakko Lehtinen.
\newblock Audio-driven facial animation by joint end-to-end learning of pose
  and emotion.
\newblock \emph{ACM Transactions on Graphics (TOG)}, 36\penalty0 (4):\penalty0
  94, 2017.

\bibitem[Tian et~al.(2019)Tian, Yuan, and Liu]{tian2019audio2face}
Guanzhong Tian, Yi~Yuan, and Yong Liu.
\newblock Audio2face: Generating speech/face animation from single audio with
  attention-based bidirectional lstm networks.
\newblock In \emph{2019 IEEE International Conference on Multimedia \& Expo
  Workshops (ICMEW)}, pages 366--371. IEEE, 2019.

\bibitem[Fried et~al.(2019)Fried, Tewari, Zollh{\"o}fer, Finkelstein,
  Shechtman, Goldman, Genova, Jin, Theobalt, and Agrawala]{fried2019text}
Ohad Fried, Ayush Tewari, Michael Zollh{\"o}fer, Adam Finkelstein, Eli
  Shechtman, Dan~B Goldman, Kyle Genova, Zeyu Jin, Christian Theobalt, and
  Maneesh Agrawala.
\newblock Text-based editing of talking-head video.
\newblock \emph{ACM Transactions on Graphics (TOG)}, 38\penalty0 (4):\penalty0
  1--14, 2019.

\bibitem[Zhou et~al.(2019)Zhou, Liu, Liu, Luo, and Wang]{zhou2019talking}
Hang Zhou, Yu~Liu, Ziwei Liu, Ping Luo, and Xiaogang Wang.
\newblock Talking face generation by adversarially disentangled audio-visual
  representation.
\newblock In \emph{Proceedings of the AAAI Conference on Artificial
  Intelligence}, volume~33, pages 9299--9306, 2019.

\bibitem[Vougioukas et~al.(2019)Vougioukas, Center, Petridis, and
  Pantic]{vougioukas2019end}
Konstantinos Vougioukas, Samsung~AI Center, Stavros Petridis, and Maja Pantic.
\newblock End-to-end speech-driven realistic facial animation with temporal
  gans.
\newblock In \emph{Proceedings of the IEEE Conference on Computer Vision and
  Pattern Recognition Workshops}, pages 37--40, 2019.

\bibitem[Zakharov et~al.(2019)Zakharov, Shysheya, Burkov, and
  Lempitsky]{zakharov2019few}
Egor Zakharov, Aliaksandra Shysheya, Egor Burkov, and Victor Lempitsky.
\newblock Few-shot adversarial learning of realistic neural talking head
  models.
\newblock \emph{arXiv preprint arXiv:1905.08233}, 2019.

\bibitem[Wang et~al.(2019)Wang, Liu, Chen, Hu, and Lian]{wang2019neural}
Zipeng Wang, Zhaoxiang Liu, Zezhou Chen, Huan Hu, and Shiguo Lian.
\newblock A neural virtual anchor synthesizer based on seq2seq and gan models.
\newblock In \emph{2019 IEEE International Symposium on Mixed and Augmented
  Reality Adjunct (ISMAR-Adjunct)}, pages 233--236. IEEE, 2019.

\bibitem[Chung et~al.(2017)Chung, Jamaludin, and Zisserman]{chung2017you}
Joon~Son Chung, Amir Jamaludin, and Andrew Zisserman.
\newblock You said that?
\newblock \emph{arXiv preprint arXiv:1705.02966}, 2017.

\bibitem[Xiao and Wang(2018)]{xiao2018dense}
Lei Xiao and Zengfu Wang.
\newblock Dense convolutional recurrent neural network for generalized speech
  animation.
\newblock In \emph{2018 24th International Conference on Pattern Recognition
  (ICPR)}, pages 633--638. IEEE, 2018.

\bibitem[Yi et~al.(2020)Yi, Ye, Zhang, Bao, and Liu]{yi2020audio}
Ran Yi, Zipeng Ye, Juyong Zhang, Hujun Bao, and Yong-Jin Liu.
\newblock Audio-driven talking face video generation with natural head pose.
\newblock \emph{arXiv preprint arXiv:2002.10137}, 2020.

\bibitem[Thies et~al.(2019)Thies, Elgharib, Tewari, Theobalt, and
  Nie{\ss}ner]{thies2019neural}
Justus Thies, Mohamed Elgharib, Ayush Tewari, Christian Theobalt, and Matthias
  Nie{\ss}ner.
\newblock Neural voice puppetry: Audio-driven facial reenactment.
\newblock \emph{arXiv preprint arXiv:1912.05566}, 2019.

\bibitem[Kumar et~al.(2017)Kumar, Sotelo, Kumar, de~Br{\'e}bisson, and
  Bengio]{kumar2017obamanet}
Rithesh Kumar, Jose Sotelo, Kundan Kumar, Alexandre de~Br{\'e}bisson, and
  Yoshua Bengio.
\newblock Obamanet: Photo-realistic lip-sync from text.
\newblock \emph{arXiv preprint arXiv:1801.01442}, 2017.

\bibitem[Bregler et~al.(1997)Bregler, Covell, and Slaney]{bregler1997video}
Christoph Bregler, Michele Covell, and Malcolm Slaney.
\newblock Video rewrite: driving visual speech with audio.
\newblock In \emph{Siggraph}, volume~97, pages 353--360, 1997.

\bibitem[Busso et~al.(2007)Busso, Deng, Grimm, Neumann, and
  Narayanan]{busso2007rigid}
Carlos Busso, Zhigang Deng, Michael Grimm, Ulrich Neumann, and Shrikanth
  Narayanan.
\newblock Rigid head motion in expressive speech animation: Analysis and
  synthesis.
\newblock \emph{IEEE Transactions on Audio, Speech, and Language Processing},
  15\penalty0 (3):\penalty0 1075--1086, 2007.

\bibitem[Taylor et~al.(2017)Taylor, Kim, Yue, Mahler, Krahe, Rodriguez,
  Hodgins, and Matthews]{taylor2017deep}
Sarah Taylor, Taehwan Kim, Yisong Yue, Moshe Mahler, James Krahe,
  Anastasio~Garcia Rodriguez, Jessica Hodgins, and Iain Matthews.
\newblock A deep learning approach for generalized speech animation.
\newblock \emph{ACM Transactions on Graphics (TOG)}, 36\penalty0 (4):\penalty0
  93, 2017.

\bibitem[Xie and Liu(2007)]{xie2007coupled}
Lei Xie and Zhi-Qiang Liu.
\newblock A coupled hmm approach to video-realistic speech animation.
\newblock \emph{Pattern Recognition}, 40\penalty0 (8):\penalty0 2325--2340,
  2007.

\bibitem[Martens and Sutskever(2011)]{martens2011learning}
James Martens and Ilya Sutskever.
\newblock Learning recurrent neural networks with hessian-free optimization.
\newblock In \emph{Proceedings of the 28th International Conference on Machine
  Learning (ICML-11)}, pages 1033--1040. Citeseer, 2011.

\bibitem[Gers et~al.(1999)Gers, Schmidhuber, and Cummins]{gers1999learning}
Felix~A Gers, J{\"u}rgen Schmidhuber, and Fred Cummins.
\newblock Learning to forget: Continual prediction with lstm.
\newblock 1999.

\bibitem[Culurciello(2018)]{culurciello2018fall}
Eugene Culurciello.
\newblock The fall of rnn/lstm.
\newblock \emph{Towards Data Science}, 2018.

\bibitem[Bai et~al.(2018)Bai, Kolter, and Koltun]{bai2018empirical}
Shaojie Bai, J~Zico Kolter, and Vladlen Koltun.
\newblock An empirical evaluation of generic convolutional and recurrent
  networks for sequence modeling.
\newblock \emph{arXiv preprint arXiv:1803.01271}, 2018.

\bibitem[Oord et~al.(2016)Oord, Dieleman, Zen, Simonyan, Vinyals, Graves,
  Kalchbrenner, Senior, and Kavukcuoglu]{oord2016wavenet}
Aaron van~den Oord, Sander Dieleman, Heiga Zen, Karen Simonyan, Oriol Vinyals,
  Alex Graves, Nal Kalchbrenner, Andrew Senior, and Koray Kavukcuoglu.
\newblock Wavenet: A generative model for raw audio.
\newblock \emph{arXiv preprint arXiv:1609.03499}, 2016.

\bibitem[Philippe(2018)]{Philippe2018}
Remy Philippe.
\newblock keras-tcn.
\newblock \url{https://github.com/philipperemy/keras-tcn}, 2018.

\bibitem[Wang et~al.(2018{\natexlab{a}})Wang, Liu, Zhu, Tao, Kautz, and
  Catanzaro]{wang2018high}
Ting-Chun Wang, Ming-Yu Liu, Jun-Yan Zhu, Andrew Tao, Jan Kautz, and Bryan
  Catanzaro.
\newblock High-resolution image synthesis and semantic manipulation with
  conditional gans.
\newblock In \emph{Proceedings of the IEEE conference on computer vision and
  pattern recognition}, pages 8798--8807, 2018{\natexlab{a}}.

\bibitem[Wang et~al.(2010)Wang, Qian, Han, and Soong]{wang2010synthesizing}
Lijuan Wang, Xiaojun Qian, Wei Han, and Frank~K Soong.
\newblock Synthesizing photo-real talking head via trajectory-guided sample
  selection.
\newblock In \emph{Eleventh Annual Conference of the International Speech
  Communication Association}, 2010.

\bibitem[Wang et~al.(2018{\natexlab{b}})Wang, Liu, Zhu, Liu, Tao, Kautz, and
  Catanzaro]{wang2018video}
Ting-Chun Wang, Ming-Yu Liu, Jun-Yan Zhu, Guilin Liu, Andrew Tao, Jan Kautz,
  and Bryan Catanzaro.
\newblock Video-to-video synthesis.
\newblock \emph{arXiv preprint arXiv:1808.06601}, 2018{\natexlab{b}}.

\bibitem[Chan et~al.(2019)Chan, Ginosar, Zhou, and Efros]{chan2019everybody}
Caroline Chan, Shiry Ginosar, Tinghui Zhou, and Alexei~A Efros.
\newblock Everybody dance now.
\newblock In \emph{Proceedings of the IEEE International Conference on Computer
  Vision}, pages 5933--5942, 2019.

\bibitem[McAllister et~al.(1997)McAllister, Rodman, Bitzer, and
  Freeman]{mcallister1997lip}
David~F McAllister, Robert~D Rodman, Donald~L Bitzer, and Andrew~S Freeman.
\newblock Lip synchronization of speech.
\newblock In \emph{Audio-Visual Speech Processing: Computational \& Cognitive
  Science Approaches}, 1997.

\bibitem[Jiang et~al.(2002)Jiang, Alwan, Keating, Auer, and
  Bernstein]{jiang2002relationship}
Jintao Jiang, Abeer Alwan, Patricia~A Keating, Edward~T Auer, and Lynne~E
  Bernstein.
\newblock On the relationship between face movements, tongue movements, and
  speech acoustics.
\newblock \emph{EURASIP Journal on Advances in Signal Processing},
  2002\penalty0 (11):\penalty0 506945, 2002.

\bibitem[Vasquez and Lewis(2019)]{vasquez2019melnet}
Sean Vasquez and Mike Lewis.
\newblock Melnet: A generative model for audio in the frequency domain.
\newblock \emph{arXiv preprint arXiv:1906.01083}, 2019.

\bibitem[Graves and Schmidhuber(2005)]{graves2005framewise}
Alex Graves and J{\"u}rgen Schmidhuber.
\newblock Framewise phoneme classification with bidirectional lstm and other
  neural network architectures.
\newblock \emph{Neural networks}, 18\penalty0 (5-6):\penalty0 602--610, 2005.

\bibitem[Isola et~al.(2016)Isola, Zhu, Zhou, and Efros]{isola2016image}
Phillip Isola, Jun-Yan Zhu, Tinghui Zhou, and Alexei~A Efros.
\newblock Image-to-image translation with conditional adversarial networks.
\newblock \emph{arXiv preprint arXiv:1611.07004}, 2016.

\end{thebibliography}


\end{document}